%% file: main_v4_camera_ready.tex
\documentclass[10pt,letterpaper]{article}
\usepackage{cogsci}
\cogscifinalcopy
\input{config}

\title{Multi-Level Narrative Evaluation Outperforms Lexical Features for Mental Health}

\author{%
    Yuxi Ma$^{1,2,4,5\,*}$, Jieming Cui$^{1,2,4,5\,*}$, Muyang Li$^{2\,*}$, Ye Zhao$^{2,6}$, Yu Li$^{2}$, Yixuan Wang$^{2}$,
    \\\vspace{3pt}%
    \normalfont\textbf{Chi Zhang$^{3,4\,\textrm{\Letter}}$, Yinyin Zang$^{2,5\,\textrm{\Letter}}$, and Yixin Zhu$^{2,1,4,5\,\textrm{\Letter}}$}
    \vspace{6pt}\\
    \small $^\star{}$Equal contributors\quad
    \small Project Website: \url{https://mayuxi.com/research/theraputic_writing}\\
    \small $^1$ Institute for Artificial Intelligence, Peking University\quad
    \small $^2$ School of Psychological and Cognitive Sciences, Peking University\\
    \small $^3$ School of Intelligence Science and Technology, Peking University\quad
    \small $^4$ State Key Laboratory of General AI, Peking University\\
    \small $^5$ Beijing Key Laboratory of Behavior and Mental Health, Peking University\quad
    \small $^6$ PKU-Changsha Institute for Computing and Digital Economy
    \vspace{-6pt}
}

\begin{document}
\maketitle

\begin{abstract}
% [Field introduction]
How people narrate their experiences offers a window into how the mind organizes them.
% [Background]
Computational approaches to therapeutic writing have evolved from lexical counting to neural methods, yet remain fragmented: dictionary tools miss discourse structure, while embeddings conflate local coherence with global organization.
% [Problem]
No existing framework maps these techniques onto the hierarchical processes through which narratives are constructed.
% [Main result]
Here we introduce a three-level framework---micro-level lexical features, meso-level semantic embeddings, and macro-level \ac{llm} narrative evaluation---and show, across 830 Chinese therapeutic texts spanning depression, anxiety, and trauma, that macro-level evaluation substantially outperforms lexical and embedding features for mental health prediction.
% [Comparison to prior knowledge]
This challenges the field's emphasis on word-counting: \emph{formal structural features} (Labov's story grammar, \ac{rst} coherence, propositional composition) demonstrate that narrative organization per se carries predictive signal, while \emph{clinically-grounded narrative dimensions} capture how psychological states are expressed through discourse. Semantic embeddings add minimal independent value but yield incremental gains in multi-level classification.
% [Broader perspective]
By grounding computational levels in discourse processing theory, this framework identifies macro-structural organization as the primary locus of clinical signal and generates testable hypotheses for intervention design and longitudinal research.

\end{abstract}

\section{Introduction}

How we narrate our lives offers a window into how the mind organizes experience. Individuals experiencing depression, anxiety, or trauma construct stories that differ from healthy narratives not merely in content, but in structural organization. While coherent, integrated life narratives predict psychological well-being \citep{adler2016incremental}, therapeutic writing interventions demonstrate clinical efficacy across diverse samples \citep{pennebaker2016opening}. Yet it remains unclear which levels of narrative construction carry the strongest mental health signal.

Current computational approaches are theoretically fragmented. Dictionary-based methods such as \ac{liwc} \citep{pennebaker2015development} capture word-level frequencies but overlook discourse structures central to therapeutic change \citep{boyd2021natural,mehl2006quantitative}. Distributional embeddings \citep{mikolov2013distributed} quantify semantic similarity yet conflate local coherence with global narrative organization. Deep neural classifiers achieve high accuracy but function as black boxes lacking interpretability \citep{teng2022survey}. This fragmentation reflects a deeper theoretical absence: no framework maps these computational techniques onto the hierarchical processes through which narratives are constructed.

\begin{figure}[t!]
    \centering
    \includegraphics[width=\linewidth]{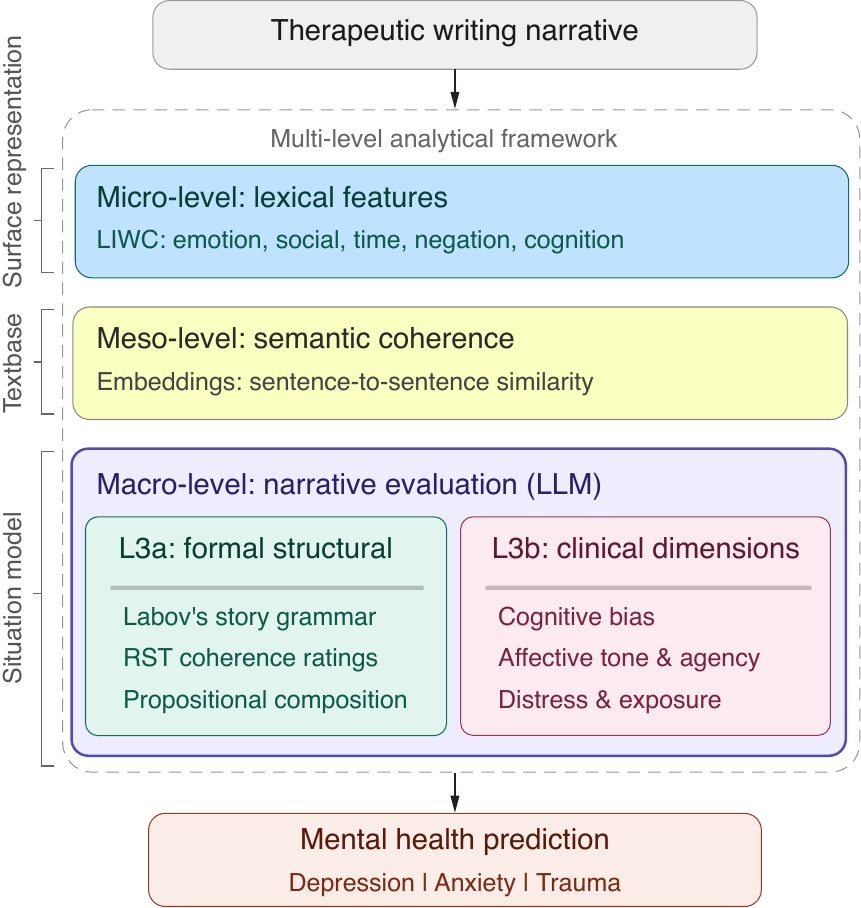}
    \caption{\textbf{Multi-level analytical framework for therapeutic writing.} Three computational layers---grounded in discourse processing theory \citep{kintsch1978toward}---operationalize hierarchical narrative construction. The micro-level captures lexical patterns via \acs{liwc}; the meso-level quantifies semantic coherence through sentence embeddings; the macro-level employs \acsp{llm} as structured evaluators, distinguishing \emph{formal structural features} (L3a: Labov's story grammar, \acs{rst} coherence, propositional composition) from \emph{clinically-grounded narrative dimensions} (L3b: cognitive bias, affective tone, distress). Left annotations map each layer onto established levels of text representation. All three layers feed into prediction of depression, anxiety, and trauma severity.}
    \label{fig:framework}
\end{figure}

Established discourse processing models provide a natural scaffold. \citet{kintsch1978toward} and subsequent work \citep{graesser2004coh,van1983strategies} propose that text comprehension and production operate across three levels: surface representations (verbatim linguistic form), textbases (propositional meaning and local coherence), and situation models (global mental simulations). Mental health conditions are thought to affect narratives across all three levels---depression is associated with ruminative, self-focused surface language \citep{rude2004language}, trauma with fragmented propositional flow \citep{brewin2010intrusive}, and both with impoverished global narrative structure \citep{adler2016incremental}. Yet no existing approach operationalizes this hierarchy to analyze therapeutic writing.

We address this gap by integrating symbolic, distributional, and generative methods into a unified framework grounded in this three-level architecture (see \cref{fig:framework}). The micro-level layer employs Chinese \ac{liwc} to capture automatic lexical choices reflecting psycho-affective states. The meso-level layer uses text embeddings to quantify semantic coherence through sentence-to-sentence propositional flow. At the macro-level, \acp{llm} function as structured evaluators extracting multi-dimensional narrative assessments. Critically, we distinguish two analytically separable sub-components within this layer: (i) \emph{formal structural features}---Labov's story grammar components, \ac{rst} coherence ratings, and compositional ratios---that capture the organizational architecture of the narrative; and (ii) \emph{clinically-grounded narrative dimensions} that assess how psychological states (cognitive bias, distress, affective tone) are organized and expressed through discourse.

Analyzing 830 Chinese therapeutic writing samples from an ecologically diverse sample spanning ages 9--50 across clinical and community settings, we find that macro-level evaluation substantially outperforms lexical and semantic features for predicting depression, anxiety, and trauma severity. We make three contributions: (i) a theoretically grounded multi-level framework that aligns computational methods with established discourse processing theory; (ii) evidence that narrative organization, not vocabulary, is the primary carrier of mental health signal in therapeutic writing; and (iii) condition-specific narrative signatures (\eg, temporal disorganization in depression, spatial grounding deficits in anxiety) that generate testable hypotheses for future intervention and longitudinal research.

\section{Related Work}

\paragraph*{Therapeutic writing and mental health}
Expressive writing about emotional experiences produces measurable improvements in psychological and physical health \citep{pennebaker2016opening}, with meta-analyses demonstrating moderate effect sizes across depression, anxiety, and trauma-related disorders \citep{frattaroli2006experimental}. The dominant tool for quantifying these shifts, \ac{liwc} \citep{pennebaker2015development}, categorizes words into psycholinguistic dimensions and has identified reliable linguistic markers---elevated negative emotion words and first-person singular pronouns in depression, increased causal and cognitive processing terms during recovery \citep{rude2004language}. However, \ac{liwc}'s reliance on fixed word lists prevents it from capturing context-dependent meanings, and its bag-of-words architecture neglects the semantic relationships and discourse structures that are central to therapeutic change \citep{taraban2019analyzing}. These limitations motivate approaches that move beyond surface-level word counting.

\paragraph*{Semantic coherence and computational analysis}
One such direction targets semantic coherence, the local integration of adjacent discourse units, serving as a clinical marker of cognitive-linguistic stability. Disorganized propositional flow and ruminative fragmentation are characteristic of both depression and trauma-related symptoms \citep{brewin2010intrusive,nolen1991responses}. Computational methods have quantified coherence through lexical overlap \citep{halliday1976cohesion}, entity grids \citep{barzilay2008modeling}, and semantic vector similarity \citep{zirikly2019clpsych}, but these features predominantly capture local, sentence-to-sentence transitions. A persistent gap remains: existing approaches often conflate such meso-level transitions with macro-level narrative organization, failing to distinguish the maintenance of local semantic flow from the global construction of narrative meaning.

\paragraph*{\acsp{llm} as structured evaluators in mental health}
\acp{llm} offer a potential path beyond this conflation. Their application in computational psychiatry has expanded rapidly, spanning suicide risk assessment, psychological consulting, and clinical text analysis \citep{yang2023towards,sharma2023human,song2025typing}. Yet most \ac{llm} applications remain end-to-end classifiers that inherit the ``black-box'' opacity of deep learning \citep{guo2024large}. An emerging alternative employs \acp{llm} not as classifiers but as structured evaluators, using prompt engineering to operationalize theoretically grounded dimensions \citep{guo2024large,stade2024large}. This zero-shot evaluative paradigm has shown promise in educational assessment \citep{mizumoto2023exploring} and cognitive bias detection \citep{ke2024mitigating}, but its potential for decoding the macro-structural organization of clinical narratives---where global coherence and rhetorical logic are paramount---remains largely unexplored.

\section{Methods}

\subsection{Dataset and Participants}

We analyzed 830 Chinese therapeutic writing samples ($>100$ words) from individuals aged 9 to 50 years ($M=20.7$, $SD=7.7$; 76.4\% female) who completed 20--30 minute expressive writing sessions about emotionally significant experiences. Samples were drawn from six therapeutic interventions conducted between 2018 and 2024, spanning clinical, post-disaster, school-based, and online settings. The clinical adult sample comprised Writing Exposure Therapy (\citealp{li2025written}; $n=30$; $M_{\text{age}}=28.6\pm7.9$; 73.3\% female) and Guided Writing Exposure (\citealp{li2025online}; $n=84$; $M_{\text{age}}=27.0\pm6.0$; 86.9\% female). Post-disaster child and adolescent samples included the Sichuan Earthquake Children study ($n=159$; $M_{\text{age}}=11.1\pm1.2$; 52.8\% female) and the Jishishan Group Intervention ($n=43$; $M_{\text{age}}=14.3\pm0.8$; 62.8\% female). Additional samples came from Hubei Primary Students preventive interventions ($n=50$; $M_{\text{age}}=10.0\pm0.2$; 74.0\% female) and Tencent Medical Platform online users ($n=464$; $M_{\text{age}}=24.0\pm5.1$; 84.3\% female). Across the pool, 137 participants (16.5\%) met criteria for clinical depression, 100 (12.0\%) for clinical anxiety, and 323 (38.9\%) for probable \ac{ptsd}.

Depression was measured using the Beck Depression Inventory-II \citep[BDI-II;][]{beck1996comparison}, Patient Health Questionnaire-9 \citep[PHQ-9;][]{kroenke2001phq}, Patient Health Questionnaire-4 \citep[PHQ-4;][]{kroenke2009ultra}, or the Revised Child Anxiety and Depression Scale \citep[RCADS-47 or RCADS-25;][]{chorpita2000assessment,ebesutani2012revised}. Anxiety was assessed via the Beck Anxiety Inventory \citep[BAI;][]{beck1988inventory}, Generalized Anxiety Disorder-7 \citep[GAD-7;][]{spitzer2006brief}, PHQ-4, or RCADS. Trauma symptoms were evaluated using the \acs{ptsd} Symptom Scale Interview for DSM-5 \citep[PSSI-5;][]{foa2016psychometric}, Child \acs{ptsd} Symptom Scale for DSM-5 -- Interview Version \citep[CPSS-5-I;][]{foa2018psychometrics}, Children's Revised Impact of Event Scale \citep[CRIES;][]{smith2003principal}, or Primary Care \acs{ptsd} Screen \citep[PC-PTSD-5;][]{prins2016primary}. Because the six sub-studies employed different instruments, raw scores were normalized to a 0--1 range by dividing by each instrument's maximum possible score. This linear rescaling preserves within-instrument rank ordering while enabling cross-study pooling, though it assumes approximate comparability of severity thresholds across instruments---a simplification we acknowledge as a limitation. Normalized scores were then converted into ordinal severity levels: four for depression and anxiety (none, mild, moderate, severe) and five for trauma (none, mild, moderate, severe, very severe). All participants provided informed consent; all contributing studies received institutional ethics approval.

\subsection{Multi-Level Analytical Framework}

\paragraph*{Layer 1: Micro-level (lexical features)}
The micro-level captures linguistic patterns through theory-driven feature selection from Simplified Chinese \ac{liwc} \citep{gao2013developing}. Each feature maps to a specific psychological mechanism: \textit{first-person singular pronouns} reflect self-focused attention \citep{pyszczynski1987self}; \textit{negative emotion words} index negative cognitive bias \citep{beck1979cognitive}; certitude and discrepancy markers (\textit{certain}, \textit{discrep}) capture absolutist thinking versus recognition of situational gaps \citep{egan2011perfectionism}; \textit{social words} indicate interpersonal engagement \citep{teo2013social}; \textit{past-tense focus} marks rumination \citep{nolen1991responses}; \textit{death-related terms} signal suicidal ideation \citep{pennebaker2015development}; and \textit{negation words} (\textit{negate}) proxy defensive cognitive operations \citep{pennebaker1997writing}.

\paragraph*{Layer 2: Meso-level (semantic coherence)}
The meso-level quantifies semantic integration using OpenAI's \texttt{text-embedding-3-small} model, generating 1536-dimensional vectors for each sentence and the complete text. \textit{Local coherence} was operationalized via sentence-to-sentence (s2s) cosine similarity, yielding indices of average propositional flow (s2s\_mean), abrupt thematic shifts (s2s\_min), and transition variability (s2s\_std). \textit{Global coherence} was captured through sentence-to-document (s2d) similarity, measuring thematic alignment (s2d\_mean), wandering (s2d\_std), and representative extremes (s2d\_max/min). Local measures reflect the fluidity of adjacent idea transitions; global measures index the narrator's ability to anchor propositions to a central theme \citep{li2014model}.

\paragraph*{Layer 3: Macro-level (narrative evaluation)}
At the macro-level, GPT-4o functions as a structured evaluator, extracting multi-dimensional narrative assessments via deterministic sampling (temperature\,=\,0) with JSON-formatted outputs and supporting textual evidence, ensuring reproducibility and external auditability. We distinguish two sub-components within this layer, operationalized through three complementary evaluation protocols.

\textbf{L3a: Formal structural features}\quad
This sub-component captures the organizational architecture of the narrative independent of clinical content, via two evaluation protocols: (i) \emph{Propositional and rhetorical logic.} Guided by macrostructure theory \citep{van2019macrostructures}, \acp{llm} decomposed narratives into minimal semantic units---each comprising a core predicate and its subject---and categorized them into Actions \& Facts, Sensory Perception, Direct Emotion, Indirect Emotion, and Cognition. These were aggregated into three functional dimensions: cognitive processing, affective engagement, and narrative grounding, capturing the allocation of conceptual resources during writing \citep{bruner1991narrative}. Concurrently, following \ac{rst} \citep{mann1988rhetorical}, \acp{llm} identified rhetorical relations (\eg, Elaboration, Cause, Contrast) and assessed transition quality on 0--5 scales, aggregated into a global coherence score. (ii) \emph{Canonical narrative organization.} \acp{llm} evaluated texts against Labov's story grammar \citep{labov1972language}, rating 6 structural components (abstract, orientation, complicating action, evaluation, resolution, and coda) on 5-point scales with supporting evidence. These scores index the narrator's capacity for coherent macrostructure and meaning-making, both robust predictors of psychological well-being \citep{adler2012living,adler2016incremental}.

\textbf{L3b: Clinically-grounded narrative dimensions}\quad
This sub-component assesses how mental health states are expressed and organized via narrative discourse. \acp{llm} evaluated 15 dimensions drawn from trauma-focused \acs{cbt} \citep{cohen2006treating} and cognitive therapy models \citep{beck1979cognitive}, organized into 4 categories: (i) \emph{structural trauma processing} examines whether the narrative macrostructure supports exposure acceptance or manifests avoidance, overgeneralization, and episodic specificity loss \citep{williams2007autobiographical}; (ii) \emph{cognitive processing} identifies how structural organization expresses cognitive biases and depth of sense-making; (iii) \emph{affective/agentic integration} measures agency and affective tone within the story arc; and (iv) \emph{global structural coherence} assesses spatio-temporal consistency and contextual density. Crucially, these dimensions capture the \emph{narrative expression} of psychological states---how cognitive patterns and affective experiences are organized within discourse---rather than direct symptom reports. For instance, \texttt{cognitive\_bias\_score} reflects how reasoning patterns structure narrative (\eg, the degree to which causal attributions, absolutist language, and negative schema are narratively organized), not a questionnaire response about thought frequency.

\subsection{Evaluation and Metrics}

We evaluated predictive utility through two complementary tasks: continuous symptom regression (depression and anxiety) and multi-class severity classification (depression, anxiety, and trauma). Trauma was excluded from regression due to the absence of continuous symptom scores in the source datasets. Regression employed \textit{ExtraTreesRegressor}; classification used \textit{Gradient Boosting} and \textit{ExtraTrees} with class-balanced weights to accommodate severity-level imbalance. All models were validated via stratified 5-fold cross-validation across 7 feature combinations---from baseline demographics (age and gender) alone to the full B+L1+L2+L3 suite---enabling systematic assessment of each layer's marginal contribution. Given minimal L3 missingness (0.4\%), zero-filling preserved cross-model comparability without introducing synthetic bias. Regression performance was assessed with $R^2$, \ac{rmse}, and \ac{mae}; classification with \ac{auc}, Balanced Accuracy, and Macro-F1.

\section{Results}

\subsection{Model Performance Across Feature Combinations}

\begin{figure}[t!]
    \centering
    \includegraphics[width=\linewidth]{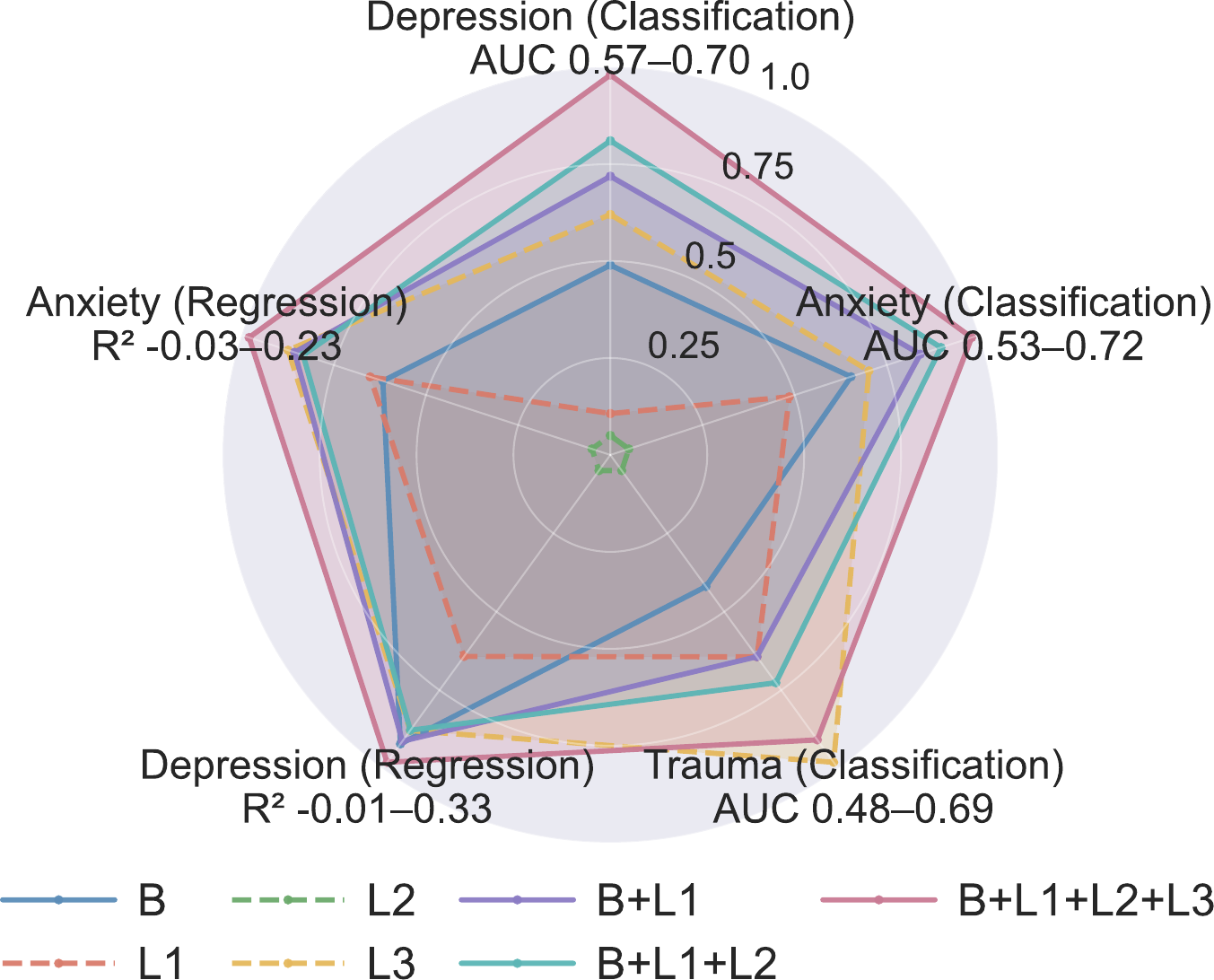}
    \caption{\textbf{Performance across layered feature sets.} Radar plot comparing regression ($R^2$) and classification (\acs{auc}) performance across five tasks. Each line represents a feature combination, from the baseline alone ($B$) through sequential layer addition up to $B+L1+L2+L3$. All metrics are normalized to a maximum of 1.00. Layer~3 (macro-level) alone approaches full-model performance, while Layer~2 (meso-level) alone falls near the plot center, indicating near-chance accuracy. Sequential integration shows incremental gains for classification but plateauing $R^2$ until Layer~3 is added.}
    \label{fig:radar}
\end{figure}

Performance varied substantially across computational layers and their combinations (\cref{tab:performance_subset,fig:radar}). The full model (B+L1+L2+L3) achieved the best performance across all tasks: depression $R^2=0.332$ and anxiety $R^2=0.235$ for regression; depression \ac{auc}=0.699, anxiety \ac{auc}=0.718, and trauma \ac{auc}=0.676 for classification.

\begin{table}[b!]
    \centering
    \small
    \setlength\tabcolsep{3pt}
    \caption{\textbf{Performance comparison across feature combinations.} $R^2$ is reported for regression (depression and anxiety); \acs{auc} for multi-class classification (depression, anxiety, and trauma). Values denote mean$\pm$SD across stratified 5-fold cross-validation. Individual layers are shown above the rule; cumulative combinations below. Bold indicates best performance per column. Trauma regression is omitted due to the absence of continuous scores in source datasets.}
    \label{tab:performance_subset}
    \resizebox{\linewidth}{!}{%
        \begin{tabular}{lccccc}
            \toprule
            \textbf{Feature Set} & \multicolumn{2}{c}{\textbf{Regression ($R^2$)$\uparrow$}} & \multicolumn{3}{c}{\textbf{Classification (\acs{auc})$\uparrow$}} \\
            \cmidrule(lr){2-3} \cmidrule(lr){4-6}
             & Dep. & Anx. & Dep. & Anx. & Trauma \\
            \midrule
            Baseline (B) & 0.311\textsuperscript{±0.115} & 0.130\textsuperscript{±0.050} & 0.629\textsuperscript{±0.017} & 0.651\textsuperscript{±0.047} & 0.567\textsuperscript{±0.040} \\
            L1 & 0.207\textsuperscript{±0.030} & 0.140\textsuperscript{±0.048} & 0.574\textsuperscript{±0.014} & 0.617\textsuperscript{±0.037} & 0.617\textsuperscript{±0.075} \\
            L2 & -0.014\textsuperscript{±0.039} & -0.034\textsuperscript{±0.019} & 0.566\textsuperscript{±0.058} & 0.527\textsuperscript{±0.058} & 0.484\textsuperscript{±0.076} \\
            L3 & 0.295\textsuperscript{±0.021} & 0.204\textsuperscript{±0.045} & 0.647\textsuperscript{±0.017} & 0.661\textsuperscript{±0.039} & 0.692\textsuperscript{±0.062} \\
            \midrule
            B + L1 & 0.308\textsuperscript{±0.027} & 0.199\textsuperscript{±0.037} & 0.662\textsuperscript{±0.014} & 0.690\textsuperscript{±0.022} & 0.617\textsuperscript{±0.064} \\
            B + L1 + L2 & 0.294\textsuperscript{±0.036} & 0.192\textsuperscript{±0.034} & 0.675\textsuperscript{±0.021} & 0.701\textsuperscript{±0.021} & 0.635\textsuperscript{±0.079} \\
            B + L1 + L2 + L3 & \textbf{0.332\textsuperscript{±0.020}} & \textbf{0.235\textsuperscript{±0.039}} & \textbf{0.699\textsuperscript{±0.018}} & \textbf{0.718\textsuperscript{±0.040}} & \textbf{0.676\textsuperscript{±0.085}} \\
            \bottomrule
        \end{tabular}%
    }%
\end{table}

When evaluated in isolation, the three layers exhibited a clear hierarchy. Layer~3 (macro-level) was the strongest single predictor, approaching full-model performance (depression $R^2=0.295$, anxiety $R^2=0.204$, trauma \ac{auc}=0.692) and substantially outperforming baseline demographics. Layer~1 (micro-level) showed moderate standalone capability comparable to the baseline. Layer~2 (meso-level) demonstrated negligible independent utility, yielding negative regression scores ($R^2=-0.014$ for depression, $-0.034$ for anxiety) and near-chance classification (\ac{auc}$\leq$0.566 across all conditions)---indicating that isolated semantic coherence features fail to capture meaningful variance beyond a mean-only baseline.

Sequential layer integration, however, revealed a more nuanced picture. Adding L1 to baseline consistently improved classification (depression \ac{auc}: 0.629$\rightarrow$0.662; anxiety: 0.651$\rightarrow$0.690; trauma: 0.567$\rightarrow$0.617), and incorporating L2 yielded further incremental gains (depression: 0.675; anxiety: 0.701; trauma: 0.635)---despite L2's poor standalone performance. This suggests that semantic coherence features capture complementary variance when combined with lexical and demographic information. The two task types showed divergent integration trajectories: classification accuracy improved incrementally with each added layer, whereas regression $R^2$ plateaued across B, B+L1, and B+L1+L2 combinations, increasing substantially only upon inclusion of Layer~3.

\begin{figure*}[t!]
    \centering
    \begin{subfigure}[t]{0.49\linewidth}
        \centering
        \includegraphics[width=\linewidth]{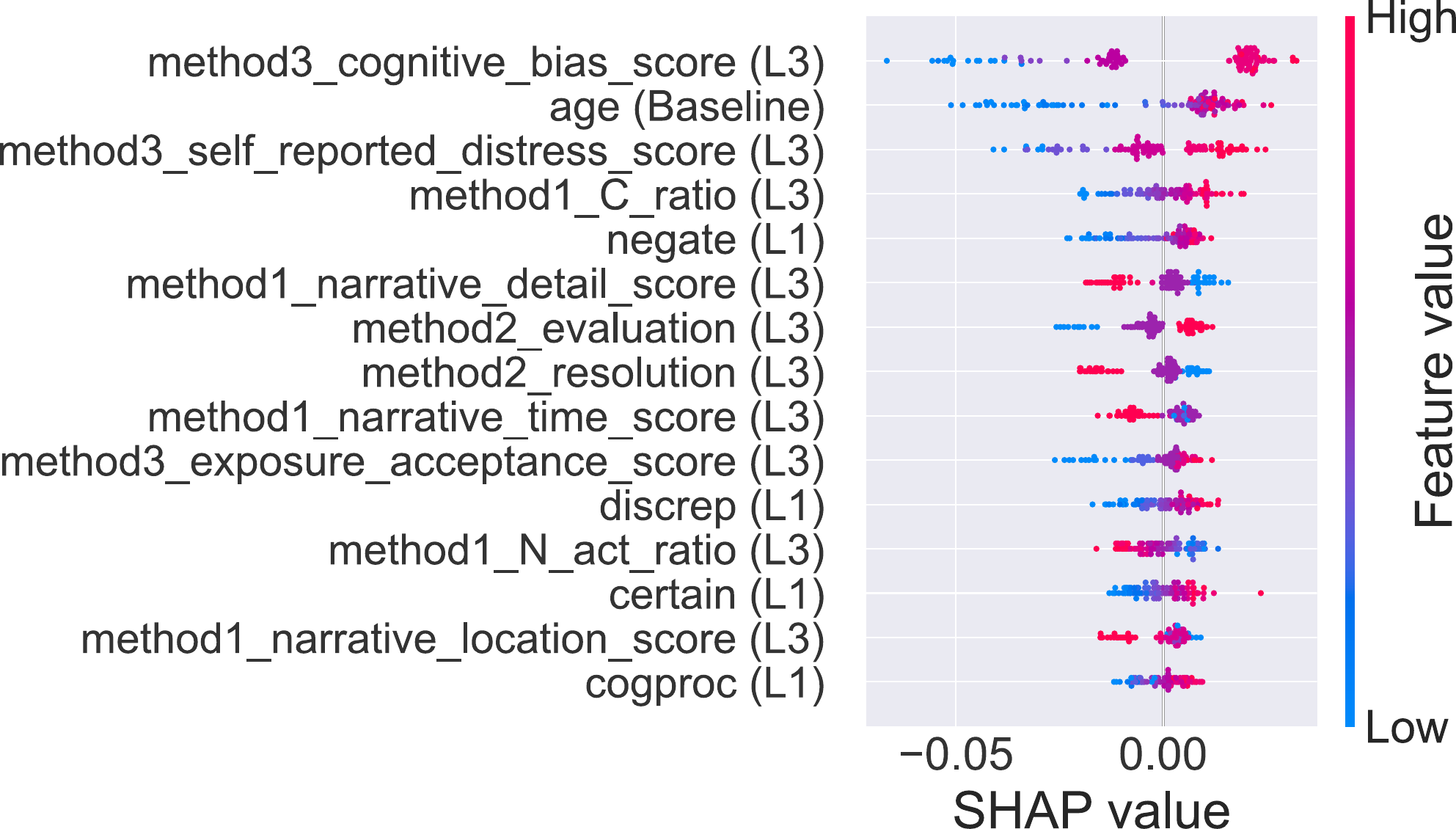}
        \caption{\textbf{Depression.} \texttt{cognitive\_bias\_score} (L3b) and age (Baseline) are the two most influential predictors. Higher cognitive bias scores drive increased predicted severity; higher Labovian \texttt{evaluation} and \texttt{resolution} scores (L3a) are associated with lower severity.}
        \label{fig:shap_depression}
    \end{subfigure}%
    \hfill
    \begin{subfigure}[t]{0.49\linewidth}
        \centering
        \includegraphics[width=\linewidth]{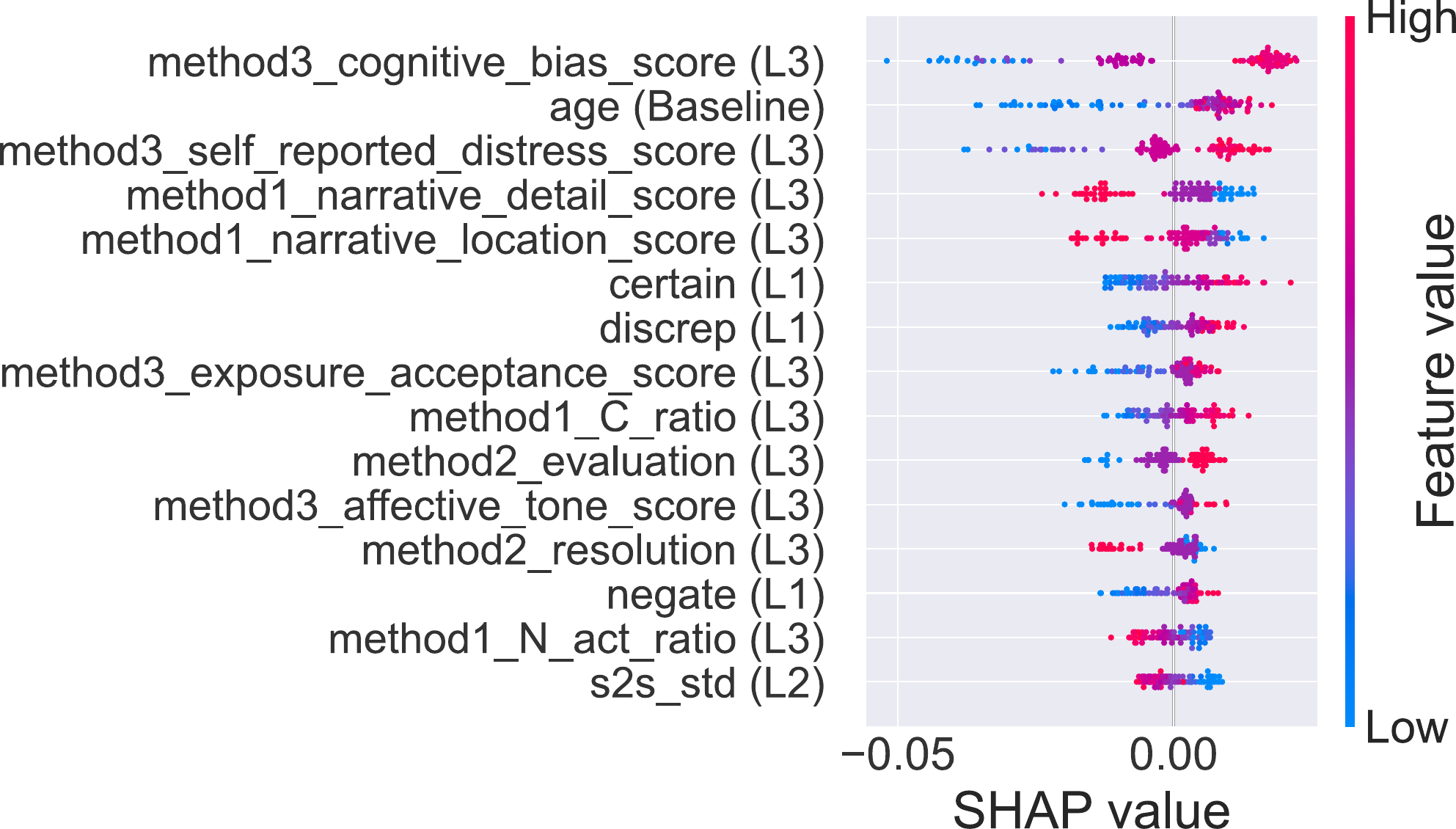}
        \caption{\textbf{Anxiety.} \texttt{cognitive\_bias\_score} (L3b) again ranks first, followed by age and \texttt{self\_reported\_distress\_score} (L3b). Anxiety-specific predictors include spatial grounding (\texttt{narrative\_location\_score}, L3a) and uncertainty markers (\texttt{certain}, \texttt{discrep}, L1).}
        \label{fig:shap_anxiety}
    \end{subfigure}%
    \caption{\textbf{\acs{shap} summary plots for depression and anxiety score prediction.} Features are ranked top-to-bottom by mean absolute \acs{shap} value (global importance). Each dot represents one sample; horizontal position indicates the \acs{shap} value (contribution to model output), and color encodes the original feature magnitude (red\,=\,high, blue\,=\,low). Feature labels indicate their computational layer: L3a (formal structural), L3b (clinically-grounded narrative dimensions), L1 (lexical), L2 (semantic coherence), or Baseline (demographics). Of the top~15 predictors, 11 are shared across both conditions, with L3 features dominating.}
    \label{fig:shap_analysis}
\end{figure*}

\subsection{Feature Importance Patterns}

\ac{shap} analysis revealed hierarchical importance patterns across the three computational layers (\cref{fig:shap_analysis}). The top~15 features for depression and anxiety showed substantial overlap (11 shared predictors), spanning L3 narrative assessments, L1 lexical markers, and baseline demographics.

Clinically-grounded narrative dimensions (L3b) ranked highest within L3. \texttt{cognitive\_bias\_score} emerged as the single strongest predictor for both depression (\ac{shap}\,=\,0.024) and anxiety (\ac{shap}\,=\,0.019), followed by \texttt{self\_reported\_distress\_score} (3rd for both). These features capture how cognitive distortions and affective distress are organized and elaborated in narrative discourse, rather than direct responses to symptom questionnaires.

Formal structural features (L3a) provided convergent evidence that narrative organization carries predictive signal independent of expressed clinical content. Labovian components (\texttt{evaluation}, \texttt{resolution}) and propositional organization scores (\texttt{narrative\_time\_score}, \texttt{C\_ratio}, \texttt{narrative\_detail\_score}) appeared consistently in the top~15 for both conditions, with a negative association with symptom severity: narratives exhibiting greater meaning-making capacity, temporal organization, and compositional balance predicted lower distress, consistent with clinical theories of narrative integration \citep{adler2012living,adler2016incremental}.

Age ranked second overall (depression \ac{shap}\,=\,0.017; anxiety \ac{shap}\,=\,0.012). Given that the six sub-samples span ages 9--50 and differ systematically in therapeutic context, age in the pooled model likely partially indexes population membership and baseline severity rather than developmental effects alone---an interpretive constraint we address in the Discussion.

Despite this shared core, condition-specific signatures emerged. Depression models favored cognitive composition (\texttt{C\_ratio}) and temporal organization (\texttt{narrative\_time\_score}), while anxiety models prioritized uncertainty markers (\texttt{certain}, \texttt{discrep}) and spatial grounding (\texttt{narrative\_location\_score}). Across both conditions, protective narrative features---structural coherence, compositional balance, narrative detail, and Labovian closure---consistently predicted lower symptom scores. Layer~2 semantic coherence features showed minimal predictive utility: all metrics fell outside the top~15 for depression, and only \texttt{s2s\_std} appeared at rank~15 for anxiety (\ac{shap}\,=\,0.004), suggesting that variability in sentence-to-sentence semantic flow may specifically characterize anxious narratives.

\section{Discussion}

This study introduced a theoretically grounded, multi-level computational framework that integrates symbolic, distributional, and generative approaches for analyzing mental health states in therapeutic writing. Across 830 Chinese narratives spanning depression, anxiety, and trauma, three key findings emerged: macro-level narrative features were the dominant predictor of symptom severity, semantic coherence contributed minimally in isolation but added value in combined models, and sequential layer integration revealed task-dependent patterns---classification benefited incrementally from each layer, while regression improved substantially only upon inclusion of macro-level structural features.

\subsection{Macro-Level Narrative Evaluation Outperforms Lexical and Embedding Approaches}

Our central finding---that macro-level narrative (L3) features explain substantially more variance than lexical or semantic features---challenges the historical emphasis on word-counting in computational psycholinguistics. This suggests that mental health status is not merely associated with the frequency of specific distress words, but is more strongly linked to the hierarchical organization of the narrator's situation model \citep{kintsch1978toward}. The performance superiority of Layer~3 indicates that \textbf{the way} a story is constructed is more informative than the specific vocabulary or textbase representations (L2 embeddings) used to tell it.

Within the macro-level layer, formal structural features (L3a) provide convergent evidence that narrative organization carries predictive signal independent of expressed content. Elements such as \texttt{narrative\_time\_score} and \texttt{narrative\_location\_score} capture the narrator's capacity to maintain a stable spatiotemporal field. Our results are consistent with the possibility that psychological distress is associated with a breakdown in this scaffolding, where high cognitive load or trauma-related dissociation may impair the construction of a coherent situational model. This pattern is particularly salient in high-severity samples, where narrative structural integrity tends to give way to disjointed propositions. These formal properties index the narrator's capacity for meaning-making closure, a robust predictor of psychological well-being \citep{adler2012living} that remains invisible to traditional lexical or embedding-based metrics.

Concurrently, the predictive power of clinically-grounded narrative dimensions (L3b) suggests that psychopathology may function as a structural filter that reshapes narrative output. Rather than appearing as isolated errors, cognitive biases (\eg, catastrophizing, all-or-nothing thinking, and over-generalization) appear to operate as systemic organizing principles that shape the narrative arc and causal reasoning. These dimensions may reflect the expansion of negative self-schema \citep{beck1979cognitive}, wherein idiosyncratic failures are codified into universal laws within the text. This suggests that \acp{llm} are not merely detecting distress markers in the text, but are capturing how distress is organized and expressed across the discourse.

Importantly, the condition-specific signatures identified through \ac{shap} analysis offer theoretically interpretable contrasts. Depression models favored cognitive composition (\texttt{C\_ratio}) and temporal organization (\texttt{narrative\_time\_score}), consistent with the ruminative temporal disorientation characteristic of depressive cognition \citep{nolen1991responses}. Anxiety models, by contrast, prioritized uncertainty markers (\textit{certain}, \textit{discrep}) and spatial grounding (\texttt{narrative\_location\_score}), aligning with the hypervigilance and environmental scanning associated with anxiety disorders. These differential patterns generate testable hypotheses: if temporal scaffolding is specifically disrupted in depression, interventions that explicitly structure narrative chronology may prove more effective for depressive symptoms than generic expressive writing.

\subsection{Methodological Contributions and the Challenge of Semantic Coherence}

This work advances computational psycholinguistics through three methodological contributions. First, we replace opaque ``black-box'' extractions with a theory-driven framework where each feature maps to specific psychological hypotheses. By operationalizing micro-level (\ac{liwc}), semantic (embeddings), and macro-structural (\ac{rst}, Labovian, \ac{cbt}) dimensions, this multi-level approach provides a transparent and interpretable link between linguistic form and psychological state.

Second, employing \acp{llm} as structured evaluators---rather than end-to-end classifiers---enhances analytical transparency. By requiring JSON-formatted outputs with supporting textual evidence, this framework ensures that Layer~3 assessments are externally auditable. This approach provides a rigorous alternative to open-ended generative assessment, bridging the gap between qualitative clinical judgment and quantitative prediction.

Third, the multi-level integration reveals a hierarchical dominance in clinical signaling: macro-structural features capture the core variance associated with psychological distress, whereas surface-level markers provide only peripheral information. This decomposition identifies the boundaries of each linguistic level, demonstrating that structural organization is a more robust correlate of psychopathology than lexical or semantic density in clinical narratives.

Notably, Layer~2's limited predictive power despite its theoretical prominence warrants reflection. While embedding-based similarity captures semantic overlap, it fails to model the sequential logic of psychological coherence. We hypothesize that in therapeutic writing, task constraints force narratives to cluster around trauma-related concepts, creating a ceiling effect for Layer~2 variance: all participants write about distressing experiences, compressing the semantic space in which embeddings operate. Layer~3, by contrast, reveals dramatic differences in \emph{how} these shared themes are structurally organized. This interpretation suggests that distributional metrics may retain diagnostic value in unconstrained corpora (\eg, social media), where topic variance is higher, but that structured clinical narratives require more granular operationalizations---such as discourse relation classifiers, argument structure analysis, or concept network graphs---to capture meaningful disruptions at the meso-level.

\subsection{Limitations and Future Directions}

Several limitations warrant consideration. First, sample heterogeneity across ages (9--50), therapeutic contexts, and assessment instruments introduces variance that age-as-covariate only partially addresses; the prominence of age in \ac{shap} rankings likely reflects sub-sample differences rather than pure developmental effects. Second, while we distinguish L3a and L3b conceptually, the current analysis reports them as a combined set; future ablation studies—coupled with validation against human expert ratings—are needed to clarify if these LLM-generated signals reflect genuine psychological constructs or redundant surface features like text length. Third, our cross-sectional design precludes causal inference: the associations between narrative disorganization and higher severity are equally consistent with distress causing fragmentation, fragmented capacity maintaining distress, or both reflecting a shared process. Longitudinal studies tracking within-person narrative evolution across treatment could illuminate causal direction. Fourth, narrative conventions and coherence markers are culturally contingent, and whether patterns in our Chinese sample generalize to typologically different languages remains an open question. Finally, translational impact requires experimental validation---the condition-specific signatures we identified suggest that scaffolding particular narrative dimensions (temporal organization for depression, spatial grounding for anxiety) could enhance therapeutic writing, a hypothesis testable through randomized controlled trials.

\paragraph{Acknowledgement}
Y. Ma, J. Cui, and Y. Zhu are supported by the National Natural Science Foundation of China (32595491, 62376009), the PKU-BingJi Joint Laboratory for Artificial Intelligence, the Wuhan Major Scientific and Technological Special Program (2025060902020304), the Hubei Embodied Intelligence Foundation Model Research and Development Program, and the National Comprehensive Experimental Base for Governance of Intelligent Society, Wuhan East Lake High-Tech Development Zone. M. Li, Y. Zhao, Y. Li, Y. Wang, and Y. Zang are supported by the National Natural Science Foundation of China (32371139, 32000776), and the Open Funding of the National Key Laboratory of Cognitive Neuroscience and Learning (CNLZD2103).

\printbibliography

\end{document}

%% file: config.tex
\usepackage[utf8]{inputenc} % allow utf-8 input
\usepackage[T1]{fontenc}    % use 8-bit T1 fonts
\usepackage{microtype}
\usepackage{times,latexsym}
\usepackage{graphicx} \graphicspath{{figures/}}
\usepackage{amsmath,nicefrac}
\usepackage{acronym}
\usepackage[breaklinks,colorlinks,citecolor=black,bookmarks=true]{hyperref}
\usepackage{balance}
\usepackage{xspace}
\usepackage{setspace}
\usepackage[skip=3pt,font=small]{subcaption}
\usepackage[skip=3pt,font=small]{caption}
\usepackage[dvipsnames,svgnames,x11names,table]{xcolor}
\usepackage[capitalise,noabbrev,nameinlink]{cleveref}
\usepackage{booktabs,tabularx,colortbl,multirow,multicol,array,makecell,tabularray}
\usepackage{enumitem}
\usepackage[misc]{ifsym}
\usepackage{dblfloatfix}
\makeatletter
\DeclareRobustCommand\onedot{\futurelet\@let@token\@onedot}
\def\@onedot{\ifx\@let@token.\else.\null\fi\xspace}
\def\eg{\emph{e.g}\onedot}

\makeatother

\makeatletter
\renewcommand{\paragraph}{%
    \@startsection{paragraph}{4}%
    {\z@}{0ex \@plus 0ex \@minus 0ex}{-1em}%
    {\normalfont\normalsize\bfseries}%
}
\makeatother

\usepackage[backend=biber,style=apa,natbib=true,annotation=false,backref=true]{biblatex}
\crefname{algorithm}{Alg.}{Algs.}
\Crefname{algocf}{Algorithm}{Algorithms}
\crefname{section}{Sec.}{Secs.}
\Crefname{section}{Section}{Sections}
\crefname{table}{Tab.}{Tabs.}
\Crefname{table}{Table}{Tables}
\crefname{figure}{Fig.}{Figs.}
\Crefname{figure}{Figure}{Figures}
\crefname{equation}{Eq.}{Eqs.}
\Crefname{equation}{Equation}{Equations}
\crefname{appendix}{Appx.}{Appxs.}
\Crefname{appendix}{Appendix}{Appendices}

\AtBeginDocument{
    \acrodef{llm}[LLM]{Large Language Model}
    \acrodef{liwc}[LIWC]{Linguistic Inquiry and Word Count}
    \acrodef{rst}[RST]{Rhetorical Structure Theory}
    \acrodef{cbt}[CBT]{Cognitive Behavioral Therapy}
    \acrodef{ptsd}[PTSD]{Post-Traumatic Stress Disorder}
    \acrodef{shap}[SHAP]{SHapley Additive exPlanations}
    \acrodef{auc}[AUC]{Area Under the Curve}
    \acrodef{rmse}[RMSE]{Root Mean Square Error}
    \acrodef{mae}[MAE]{Mean Absolute Error}
}

%% file: reference.bib
@article{adler2016incremental,
  title={The incremental validity of narrative identity in predicting well-being: A review of the field and recommendations for the future},
  author={Adler, Jonathan M and Lodi-Smith, Jennifer and Philippe, Frederick L and Houle, Iliane},
  journal={Personality and Social Psychology Review},
  volume={20},
  number={2},
  pages={142--175},
  year={2016},
  publisher={Sage Publications Sage CA: Los Angeles, CA}
}

@article{barzilay2008modeling,
  title={Modeling local coherence: An entity-based approach},
  author={Barzilay, Regina and Lapata, Mirella},
  journal={Computational Linguistics},
  volume={34},
  number={1},
  pages={1--34},
  year={2008},
  publisher={MIT Press}
}

@book{beck1979cognitive,
  title={Cognitive therapy of depression},
  author={Beck, Aaron T and Rush, A John and Shaw, Brian F and Emery, Gary and DeRubeis, Robert J and Hollon, Steven D},
  publisher={Guilford Press},
  year={1979}
}

@article{brewin2010intrusive,
  title={Intrusive images in psychological disorders: characteristics, neural mechanisms, and treatment implications.},
  author={Brewin, Chris R and Gregory, James D and Lipton, Michelle and Burgess, Neil},
  journal={Psychological Review},
  volume={117},
  number={1},
  pages={210},
  year={2010},
  publisher={American Psychological Association}
}

@article{frattaroli2006experimental,
  title={Experimental disclosure and its moderators: a meta-analysis.},
  author={Frattaroli, Joanne},
  journal={Psychological Bulletin},
  volume={132},
  number={6},
  pages={823},
  year={2006},
  publisher={American Psychological Association}
}

@inproceedings{gao2013developing,
  title={Developing simplified Chinese psychological linguistic analysis dictionary for microblog},
  author={Gao, Rui and Hao, Bibo and Li, He and Gao, Yusong and Zhu, Tingshao},
  booktitle={International Conference on Brain and Health Informatics},
  year={2013}
}

@article{graesser2004coh,
  title={Coh-Metrix: Analysis of text on cohesion and language},
  author={Graesser, Arthur C and McNamara, Danielle S and Louwerse, Max M and Cai, Zhiqiang},
  journal={Behavior Research Methods, Instruments, \& Computers},
  volume={36},
  number={2},
  pages={193--202},
  year={2004},
  publisher={Springer}
}

@book{halliday1976cohesion,
  title={Cohesion in English},
  author={Halliday, Michael Alexander Kirkwood and Hasan, Ruqaiya},
  year={1976},
  publisher={Routledge}
}

@book{labov1972language,
  title={Language in the inner city: Studies in the Black English vernacular},
  author={Labov, William},
  volume={3},
  year={1972},
  publisher={University of Pennsylvania Press}
}

@inproceedings{li2014model,
  title={A model of coherence based on distributed sentence representation},
  author={Li, Jiwei and Hovy, Eduard},
  booktitle=EMNLP,
  year={2014}
}

@article{mann1988rhetorical,
  title={Rhetorical structure theory: Toward a functional theory of text organization},
  author={Mann, William C and Thompson, Sandra A},
  journal={Text-interdisciplinary Journal for the Study of Discourse},
  volume={8},
  number={3},
  pages={243--281},
  year={1988},
  publisher={Walter de Gruyter, Berlin/New York Berlin, New York}
}

@article{mizumoto2023exploring,
  title={Exploring the potential of using an AI language model for automated essay scoring},
  author={Mizumoto, Atsushi and Eguchi, Masaki},
  journal={Research Methods in Applied Linguistics},
  volume={2},
  number={2},
  pages={100050},
  year={2023},
  publisher={Elsevier}
}

@article{nolen1991responses,
  title={Responses to depression and their effects on the duration of depressive episodes.},
  author={Nolen-Hoeksema, Susan},
  journal={Journal of Abnormal Psychology},
  volume={100},
  number={4},
  pages={569},
  year={1991},
  publisher={American Psychological Association}
}

@article{pennebaker2015development,
  title={The development and psychometric properties of LIWC2015},
  author={Pennebaker, James W and Boyd, Ryan L and Jordan, Kayla and Blackburn, Kate},
  year={2015}
}

@book{pennebaker2016opening,
  title={Opening Up by Writing It Down: How Expressive Writing Improves Health and Eases Emotional Pain},
  author={Pennebaker, James W},
  year={2016},
  publisher={Guilford Publications}
}

@article{rude2004language,
  title={Language use of depressed and depression-vulnerable college students},
  author={Rude, Stephanie and Gortner, Eva-Maria and Pennebaker, James},
  journal={Cognition \& Emotion},
  volume={18},
  number={8},
  pages={1121--1133},
  year={2004},
  publisher={Taylor \& Francis}
}

@article{sharma2023human,
  title={Human--AI collaboration enables more empathic conversations in text-based peer-to-peer mental health support},
  author={Sharma, Ashish and Lin, Inna W and Miner, Adam S and Atkins, David C and Althoff, Tim},
  journal={Nature Machine Intelligence},
  volume={5},
  number={1},
  pages={46--57},
  year={2023},
  publisher={Nature Publishing Group UK London}
}

@inproceedings{yang2023towards,
  title={Towards Interpretable Mental Health Analysis with Large Language Models},
  author={Yang, Kailai and Ji, Shaoxiong and Zhang, Tianlin and Xie, Qianqian and Kuang, Ziyan and Ananiadou, Sophia},
  booktitle=EMNLP,
  year={2023}
}

@inproceedings{zirikly2019clpsych,
  title={CLPsych 2019 shared task: Predicting the degree of suicide risk in Reddit posts},
  author={Zirikly, Ayah and Resnik, Philip and Uzuner, Ozlem and Hollingshead, Kristy},
  booktitle={The Sixth Workshop on Computational Linguistics and Clinical Psychology},
  year={2019}
}

@article{kintsch1978toward,
  title={Toward a model of text comprehension and production.},
  author={Kintsch, Walter and Van Dijk, Teun A},
  journal={Psychological Review},
  volume={85},
  number={5},
  pages={363},
  year={1978},
  publisher={American Psychological Association}
}

@article{williams2007autobiographical,
  title={Autobiographical memory specificity and emotional disorder.},
  author={Williams, J Mark G and Barnhofer, Thorsten and Crane, Catherine and Herman, Dirk and Raes, Filip and Watkins, Ed and Dalgleish, Tim},
  journal={Psychological bulletin},
  volume={133},
  number={1},
  pages={122},
  year={2007},
  publisher={American Psychological Association}
}

@article{pyszczynski1987self,
  title={Self-regulatory perseveration and the depressive self-focusing style: a self-awareness theory of reactive depression.},
  author={Pyszczynski, Tom and Greenberg, Jeff},
  journal={Psychological Bulletin},
  volume={102},
  number={1},
  pages={122},
  year={1987},
  publisher={American Psychological Association}
}

@article{egan2011perfectionism,
  title={Perfectionism as a transdiagnostic process: A clinical review},
  author={Egan, Sarah J and Wade, Tracey D and Shafran, Roz},
  journal={Clinical Psychology Review},
  volume={31},
  number={2},
  pages={203--212},
  year={2011},
  publisher={Elsevier}
}

@article{pennebaker1997writing,
  title={Writing about emotional experiences as a therapeutic process},
  author={Pennebaker, James W},
  journal={Psychological Science},
  volume={8},
  number={3},
  pages={162--166},
  year={1997},
  publisher={SAGE Publications Sage CA: Los Angeles, CA}
}

@article{adler2012living,
  title={Living into the story: agency and coherence in a longitudinal study of narrative identity development and mental health over the course of psychotherapy.},
  author={Adler, Jonathan M},
  journal={Journal of Personality and Social Psychology},
  volume={102},
  number={2},
  pages={367},
  year={2012},
  publisher={American Psychological Association}
}

@inproceedings{mikolov2013distributed,
  title={Distributed representations of words and phrases and their compositionality},
  author={Mikolov, Tomas and Sutskever, Ilya and Chen, Kai and Corrado, Greg S and Dean, Jeff},
  booktitle=NIPS,
  year={2013}
}

@article{boyd2021natural,
  title={Natural language analysis and the psychology of verbal behavior: The past, present, and future states of the field},
  author={Boyd, Ryan L and Schwartz, H Andrew},
  journal={Journal of Language and Social Psychology},
  volume={40},
  number={1},
  pages={21--41},
  year={2021},
  publisher={Sage Publications Sage CA: Los Angeles, CA}
}

@book{mehl2006quantitative,
  title={Handbook of multimethod measurement in psychology.},
  author={Eid, Michael Ed and Diener, Ed Ed},
  year={2006},
  publisher={American Psychological Association}
}

@article{teng2022survey,
  title={A survey on the interpretability of deep learning in medical diagnosis},
  author={Teng, Qiaoying and Liu, Zhe and Song, Yuqing and Han, Kai and Lu, Yang},
  journal={Multimedia Systems},
  volume={28},
  number={6},
  pages={2335--2355},
  year={2022},
  publisher={Springer}
}

@article{teo2013social,
  title={Social relationships and depression: ten-year follow-up from a nationally representative study},
  author={Teo, Alan R and Choi, HwaJung and Valenstein, Marcia},
  journal={PloS one},
  volume={8},
  number={4},
  pages={e62396},
  year={2013},
  publisher={Public Library of Science San Francisco, USA}
}

@book{van2019macrostructures,
  title={Macrostructures: An interdisciplinary study of global structures in discourse, interaction, and cognition},
  author={Van Dijk, Teun A},
  year={2019},
  publisher={Routledge}
}

@article{bruner1991narrative,
  title={The narrative construction of reality},
  author={Bruner, Jerome},
  journal={Critical inquiry},
  volume={18},
  number={1},
  pages={1--21},
  year={1991},
  publisher={University of Chicago Press}
}

@article{cohen2006treating,
  title={Treating Trauma and Traumatic Grief in Children and Adolescents},
  author={Cohen, Judith and Mannarino, Anthony and Deblinger, Esther and others},
  journal={Guilford Publications},
  year={2006},
  publisher={ERIC}
}

@article{li2025written,
  title={Written exposure therapy for posttraumatic stress disorder and integration of a mindfulness based app in China: a pilot randomized controlled trial},
  author={Li, Muyang and Zhao, Ye and Guo, Zeyu and Wei, Mingcen and Fan, Shijia and Chen, Qiang and Li, Yu and Zang, Yinyin},
  journal={Behavior Therapy},
  year={2025},
  publisher={Elsevier}
}

@article{li2025online,
  title={An Online Guided Written Exposure Therapy for Symptoms of Posttraumatic Stress Disorder: A Randomized Controlled Trial},
  author={Li, Muyang and Zhao, Ye and Rosenfield, David and Guo, Zeyu and Wei, Mingcen and Fan, Shijia and Li, Yu and Zang, Yinyin},
  journal={Psychotherapy and Psychosomatics},
  year={2025}
}

@article{beck1996comparison,
  title={Comparison of Beck Depression Inventories-IA and-II in psychiatric outpatients},
  author={Beck, Aaron T and Steer, Robert A and Ball, Roberta and Ranieri, William F},
  journal={Journal of Personality Assessment},
  volume={67},
  number={3},
  pages={588--597},
  year={1996},
  publisher={Taylor \& Francis}
}

@article{kroenke2001phq,
  title={The PHQ-9: validity of a brief depression severity measure},
  author={Kroenke, Kurt and Spitzer, Robert L and Williams, Janet BW},
  journal={Journal of General Internal Medicine},
  volume={16},
  number={9},
  pages={606--613},
  year={2001},
  publisher={Wiley Online Library}
}

@article{kroenke2009ultra,
  title={An ultra-brief screening scale for anxiety and depression: the PHQ--4},
  author={Kroenke, Kurt and Spitzer, Robert L and Williams, Janet BW and L{\"o}we, Bernd},
  journal={Psychosomatics},
  volume={50},
  number={6},
  pages={613--621},
  year={2009},
  publisher={Elsevier}
}

@article{chorpita2000assessment,
  title={Assessment of symptoms of DSM-IV anxiety and depression in children: A revised child anxiety and depression scale},
  author={Chorpita, Bruce F and Yim, Letitia and Moffitt, Catherine and Umemoto, Lori A and Francis, Sarah E},
  journal={Behaviour Research and Therapy},
  volume={38},
  number={8},
  pages={835--855},
  year={2000},
  publisher={Elsevier}
}

@article{ebesutani2012revised,
  title={The Revised Child Anxiety and Depression Scale-Short Version: scale reduction via exploratory bifactor modeling of the broad anxiety factor.},
  author={Ebesutani, Chad and Reise, Steven P and Chorpita, Bruce F and Ale, Chelsea and Regan, Jennifer and Young, John and Higa-McMillan, Charmaine and Weisz, John R},
  journal={Psychological Assessment},
  volume={24},
  number={4},
  pages={833},
  year={2012},
  publisher={American Psychological Association}
}

@article{spitzer2006brief,
  title={A brief measure for assessing generalized anxiety disorder: the GAD-7},
  author={Spitzer, Robert L and Kroenke, Kurt and Williams, Janet BW and L{\"o}we, Bernd},
  journal={Archives of Internal Medicine},
  volume={166},
  number={10},
  pages={1092--1097},
  year={2006},
  publisher={American Medical Association}
}

@article{foa2016psychometric,
  title={Psychometric properties of the Posttraumatic Stress Disorder Symptom Scale Interview for DSM--5 (PSSI--5).},
  author={Foa, Edna B and McLean, Carmen P and Zang, Yinyin and Zhong, Jody and Rauch, Sheila and Porter, Katherine and Knowles, Kelly and Powers, Mark B and Kauffman, Brooke Y},
  journal={Psychological Assessment},
  volume={28},
  number={10},
  pages={1159},
  year={2016},
  publisher={American Psychological Association}
}

@article{foa2018psychometrics,
  title={Psychometrics of the child PTSD symptom scale for DSM-5 for trauma-exposed children and adolescents},
  author={Foa, Edna B and Asnaani, Anu and Zang, Yinyin and Capaldi, Sandra and Yeh, Rebecca},
  journal={Journal of Clinical Child \& Adolescent Psychology},
  volume={47},
  number={1},
  pages={38--46},
  year={2018},
  publisher={Taylor \& Francis}
}

@article{smith2003principal,
  title={Principal components analysis of the impact of event scale with children in war},
  author={Smith, Patrick and Perrin, Sean and Dyregrov, Atle and Yule, William},
  journal={Personality and Individual Differences},
  volume={34},
  number={2},
  pages={315--322},
  year={2003},
  publisher={Elsevier}
}

@article{prins2016primary,
  title={The primary care PTSD screen for DSM-5 (PC-PTSD-5): development and evaluation within a veteran primary care sample},
  author={Prins, Annabel and Bovin, Michelle J and Smolenski, Derek J and Marx, Brian P and Kimerling, Rachel and Jenkins-Guarnieri, Michael A and Kaloupek, Danny G and Schnurr, Paula P and Kaiser, Anica Pless and Leyva, Yani E and others},
  journal={Journal of General Internal Medicine},
  volume={31},
  number={10},
  pages={1206--1211},
  year={2016},
  publisher={Springer}
}

@article{guo2024large,
  title={Large language models for mental health applications: systematic review},
  author={Guo, Zhijun and Lai, Alvina and Thygesen, Johan H and Farrington, Joseph and Keen, Thomas and Li, Kezhi and others},
  journal={JMIR Mental Health},
  volume={11},
  number={1},
  pages={e57400},
  year={2024},
  publisher={JMIR Publications Inc., Toronto, Canada}
}

@inproceedings{song2025typing,
  title={The typing cure: Experiences with large language model chatbots for mental health support},
  author={Song, Inhwa and Pendse, Sachin R and Kumar, Neha and De Choudhury, Munmun},
  booktitle=CHI,
  year={2025},
}

@article{ke2024mitigating,
  title={Mitigating cognitive biases in clinical decision-making through multi-agent conversations using large language models: simulation study},
  author={Ke, Yuhe and Yang, Rui and Lie, Sui An and Lim, Taylor Xin Yi and Ning, Yilin and Li, Irene and Abdullah, Hairil Rizal and Ting, Daniel Shu Wei and Liu, Nan},
  journal={Journal of Medical Internet Research},
  volume={26},
  pages={e59439},
  year={2024},
  publisher={JMIR Publications Toronto, Canada}
}

@book{van1983strategies,
  title={Strategies of discourse comprehension},
  author={Van Dijk, Teun Adrianus and Kintsch, Walter and others},
  year={1983},
  publisher={Academic press New York}
}

@article{taraban2019analyzing,
  title={Analyzing topic differences, writing quality, and rhetorical context in college students’ essays using linguistic inquiry and word count (liwc)},
  author={Taraban, Roman and Abusal, Khaleel},
  year={2019},
  journal={East European Journal of Psycholinguistics},
  publisher={Lesya Ukrainka Eastern European National University}
}

@article{stade2024large,
  title={Large language models could change the future of behavioral healthcare: a proposal for responsible development and evaluation},
  author={Stade, Elizabeth C and Stirman, Shannon Wiltsey and Ungar, Lyle H and Boland, Cody L and Schwartz, H Andrew and Yaden, David B and Sedoc, Jo{\~a}o and DeRubeis, Robert J and Willer, Robb and Eichstaedt, Johannes C},
  journal={NPJ Mental Health Research},
  volume={3},
  number={1},
  pages={12},
  year={2024},
  publisher={Nature Publishing Group UK London}
}

@article{beck1988inventory,
  title={An inventory for measuring clinical anxiety: psychometric properties.},
  author={Beck, Aaron T and Epstein, Norman and Brown, Gary and Steer, Robert A},
  journal={Journal of Consulting and Clinical Psychology},
  volume={56},
  number={6},
  pages={893},
  year={1988},
  publisher={American Psychological Association}
}


%% file: reference_header.bib
@string {NIPS = "{Proceedings of Advances in Neural Information Processing Systems (NeurIPS)}"}

@string {ROMAN = "{IEEE International Symposium on Robot and Human Interactive Communication (RO-MAN)}"}

@string {EMNLP = "{Annual Conference on Empirical Methods in Natural Language Processing (EMNLP)}"}

@string {CHI = "{ACM Conference on Human Factors in Computing Systems (CHI)}"}
